\documentclass[11pt]{article}
\usepackage[utf8]{inputenc}
\usepackage{amsmath, amssymb}
\usepackage[a4paper, margin=1in]{geometry}
\usepackage{setspace}
\usepackage{newpxtext}
\usepackage{authblk}
\usepackage[hyperindex,breaklinks]{hyperref}
\usepackage{xcolor}

\title{Apriori Knowledge in an Era of Computational Opacity: The Role of AI in Mathematical Discovery}

\author[1,2]{Eamon Duede\thanks{Email: eduede@purdue.edu}}
\author[3]{Kevin Davey\thanks{Email: kjdavey@uchicago.edu}}

\affil[1]{Purdue University}
\affil[2]{Argonne National Laboratory}
\affil[3]{University of Chicago}

\date{}

\begin{document}

\maketitle

\begin{abstract}
Can we acquire apriori knowledge of mathematical facts from the outputs of computer programs? People like Burge have argued (correctly in our opinion) that, for example, Appel and Haken acquired apriori knowledge of the Four Color Theorem from their computer program insofar as their program simply automated human forms of mathematical reasoning. However, unlike such programs, we argue that the opacity of modern LLMs and DNNs creates obstacles in obtaining apriori mathematical knowledge from them in similar ways. We claim though that if a proof-checker automating human forms of proof-checking is attached to such machines, then we can obtain apriori mathematical knowledge from them after all, even though the original machines are entirely opaque to us and the proofs they output may not, themselves, be human-surveyable.\\

\noindent ** This paper is forthcoming in \textit{Philosophy of Science}. Final contents are subject to potential minor revision.**
\end{abstract}

\section{Introduction}
\label{sec:intro}

The main issue this paper revolves around is what role computers can play in expanding our purely rational capacities and purely rational knowledge. When we learn a fact from a computer, should we think of ourselves as merely having done an experiment of sorts and, thus, at best, having acquired a piece of empirical knowledge? Or can learning something from a computer sometimes expand our non-empirical knowledge -- that is, can it sometimes expand our purely rational or apriori knowledge?

The obvious case to focus on here is the case of mathematics. Suppose a computer tells us that some mathematical claim is true. In the right circumstances, do we then know that mathematical fact on purely rational – that is, apriori grounds? Obviously, the answer to that question is going to depend on the details of what sort of computer we are talking about. So we will phrase the question with which we are concerned as follows:

\begin{quote}
\textbf{Main Question}: Are there situations  in which we can acquire apriori knowledge of a mathematical fact $X$ purely on the basis of a computer outputting the claim that $X$ is true? If so, what sorts of situations are these?
\end{quote}

This question concerns the acquisition of apriori knowledge \textit{purely} on the basis of the computer outputting the claim that something is true. To make clear what is being asked here, imagine that a computer \textit{both} outputs the claim that some mathematical fact $X$ is true, as well as a correct proof of $X$. If you think that in the right circumstances, merely witnessing the computer output the claim that $X$ is true gives us apriori knowledge of $X$, then you think that in such circumstances, we can acquire apriori knowledge purely on the basis of the computer outputting the claim that $X$ is true. But if you think, for example, that it is only once we check the proof of $X$ ourselves that we acquire apriori knowledge of $X$, then you do \textit{not} think that in those circumstances, we come to know $X$ \textit{purely} on the basis of the computer outputting the claim that $X$ is true.

Early discussions of our Main Question were motivated by Appel and Haken's famous 1977 computer proof of the Four Color Theorem \cite{appel1989every}, which is so long that it cannot be human-checked. Because of this, some thought that the idea that all mathematical knowledge is essentially apriori had to be rejected and room created for merely empirical or experimental mathematical knowledge.
 
However, following the work of Burge \cite{burge1998computer}, we argue in Section~\ref{sec:2} that so long as the running of a computer program can be understood as a mechanized exercise of something like ordinary human mathematical capacities, the output of a program can indeed give us apriori mathematical knowledge. We thus follow Burge in claiming that Appel and Haken did indeed acquire apriori knowledge of the truth of the Four Color Theorem from the output of their computer program. This means that our Main Question can be answered affirmatively in the case of the situation faced by Appel and Haken in 1977.

The problem, however, is that the argument of Section~\ref{sec:2} does not apply to the output of machines like deep neural networks (henceforth DNNs) and generative language learning models (henceforth LLMs), whose inner workings are, in a sense, opaque to us but are, nevertheless, of increasing value to mathematicians. In Section~\ref{sec:3}, we argue that outside special cases, we cannot \textit{directly} acquire apriori mathematical knowledge from the reports of DNNs or LLMs. A result of this kind seems to impose a strong limitation on our ability to acquire genuine mathematical knowledge from AI.

Nevertheless, in Section~\ref{sec:4}, we argue that mathematicians can overcome this limitation by applying a transparent proof-checker to an appropriately structured output of a DNN or LLM. So long as this proof-checker may be understood as a mechanized exercise of human proof-checking capacities, we claim that we can acquire genuine mathematical knowledge using opaque DNNs or LLMs from the output of the proof-checker, even though this knowledge may not be obtained directly from the DNN or LLM itself. 

In this way, we arrive at the perhaps surprising result that it is possible to acquire genuine apriori knowledge of a mathematical fact $X$ purely on the basis of the output of a computer, where a proof of $X$ has been generated by a process that is entirely opaque to us, \textit{and} is so complex that the proof is not human-checkable in any way. This suggests that AI can indeed play a significant and potentially transformative role in generating genuine mathematical knowledge and that there is perhaps a larger set of cases than one might expect in which we can answer our Main Question affirmatively and
acquire apriori knowledge purely on the basis of the output of a computer.

\section{Knowledge of the Four Color Theorem}
\label{sec:2}

Discussion of our Main Question began in earnest when, in 1977, Appel and Haken used a computer to verify the Four Color Theorem (henceforth 4CT) \cite{appel1989every}. Appel and Haken argued that to prove the 4CT, it sufficed to verify the 4-colorability of a particular set of 1,834 finite graphs. While this verification can be done entirely mechanically, for larger graphs, it is extremely time-consuming. Appel and Haken thus used a supercomputer to verify that each of these 1,834 graphs were indeed 4-colorable, requiring over a month of continual computer operation. As a result, the successful execution of their algorithm can be understood as giving a proof of the 4CT. The sheer length of this proof means, however, that it cannot be surveyed and checked step-by-step by a human mathematician. It remains true today that no human has produced a proof of the 4CT that can be checked by a human mathematician without the aid of a computer.

In 1979, philosophers began to reflect on what this meant for mathematics. Tymoczko \cite{tymoczko1979four} argued that as a result, mathematics had now become an empirical discipline in which proofs could be obtained by performing \textit{experiments} such as the running of computer programs.\footnote{Detlefsen and Luker \cite{detlefsen1980four} further argued that mathematics had in some sense always been an empirical discipline and that there was therefore nothing philosophically new in Appel and Haken's accomplishment.}

In particular, Tymoczko thought that our knowledge of the 4CT rested on an argument involving the premise
\begin{quote}
\textbf{Rel}: Carefully written computer programs reliably output true claims.
\end{quote}
Tymoczko thought of \textbf{Rel} as a claim stating the reliability of a piece of scientific equipment. It is thus not the sort of thing that can be established by pure reason. Thus, any knowledge obtained with the use of \textbf{Rel} could, at most, be empirical. Tymoczko concluded that our knowledge of the 4CT, while genuine knowledge, was not apriori -- that is, not justified purely on rational grounds -- but rather merely empirical. Others, such as \cite{detlefsen1980four}, concurred.

For reasons that we will explain, we do not find this view compelling. Instead, we are more persuaded by a different way of looking at things due to Burge \cite{burge1998computer}. Rejecting the views just described, Burge argued that the output of Appel and Haken's program gives us an apriori\footnote{In calling a justification or warrant \textit{apriori}, Burge means that it does not depend upon empirical considerations in any way for its force. Although the question of how to precisely characterize the apriori is vexed (see for example \cite{Williamson2013}), in what follows, we only need to rely on the fact that traditional and ordinary forms of mathematical argument are apriori, and will not need to posit anything controversial about the nature of the apriori.} (and not merely empirical) warrant for believing the 4CT.\footnote{Strictly speaking, the output of Appel and Haken's program only assures us that the given 1,834 graphs are 4-colorable, and to get the Four Color Theorem from this, we need to supplement it with a further piece of human-generated mathematics. For the sake of brevity, however, we shall be slightly sloppy and talk of the output of the program as giving us an apriori warrant for believing the Four Color Theorem, even though technically, we (and Burge) should only claim that it gives us an apriori warrant for believing that the given 1,834 graphs are 4-colorable.}

\subsection{Memory as a Rational Resource}

To explain why Burge thought this, it will be useful to draw a comparison with the use of human memory in mathematics. In the process of surveying or creating a mathematical proof, I typically must use my memory. When proving a theorem, there might come a point where I wonder whether I have already proven some particular lemma. I pause, recall that I have, and then continue reasoning. But in that pause, when I ask myself, ‘Have I already proven this lemma?’ and decide upon consulting my memory that, in fact, I have, does it make sense to suppose that I am doing an experiment with my brain? Is the resulting knowledge merely empirical? Can I thus only rely on my memory if I have empirical knowledge of the general reliability of my memory, for example?

Burge thinks that relying on our memory is not doing an experiment which thereby yields at most empirical knowledge. His view is rather that we have \textit{defeasible, apriori} grounds for believing what we seemingly remember. So when I have a seeming memory of something (such as proving a lemma), I am entitled on purely rational grounds to believe that I have proved the lemma. It is not the case that I must first do some memory tests to establish the reliability of my memory before I have grounds to believe what I seemingly remember. Rather, Burge's view is that I have purely rational grounds for believing the lemma, \textit{because I have a memory of proving it}. However, these grounds are, of course, defeasible because I might later realize that I was misremembering. Purely rational grounds are not infallible on this sort of picture.

To be sure, searching my memory for an episode of proving a certain lemma is something like an empirical investigation. It is only an empirical fact that I proved the lemma this morning on my whiteboard, and it is only an empirical fact that I have a memory of such a thing happening. Nevertheless, upon finding such a memory, I have purely rational (i.e., apriori) grounds for believing the lemma. The key here is to note that I do not infer the lemma from the existence of the memory, for otherwise, mathematical papers would need to be full of claims about the memories of their authors, and their results would indeed only be empirically known. Instead, I infer the lemma simply from the reasoning that has been remembered. 

More generally, then, Burge claims that, even when only exercising our rational capacities, there are various sorts of resources on which we can rely. He calls these \textit{rational resources}. When these rational resources offer us some sort of claim, we are defeasibly, apriori entitled to believe it. So, for example, my memory is a rational resource that I can draw on in my reasoning. So is the use of my visual system insofar as I am doing something like reading a proof or my notes. A thermometer, however, is not a rational resource because when I trust a thermometer, I am doing something that goes beyond mere reasoning. (A thermometer is instead an \textit{empirical} resource.) Burge's general claim is that ’\textit{resources for rationality are, other things equal, to be believed}’ \cite[p.5]{burge1998computer}. This extends to the use of these resources even outside the case of pure reasoning, though we shall not discuss the delicate details of this here. 

\subsection{Computers as Rational Resources}

With this in mind, let us return to Appel and Haken’s computer program. This program is designed to go through all the 1,834 basic maps and verify their 4-colorability in exactly the way Appel and Haken might. It organizes the basic maps systematically into a list and does exactly what they might do to check each case, though far more quickly and indefatigably. Appel and Haken’s computer is, thus, simply a mechanized application of their ordinary human rational capacities that performs their reasoning for them. Moreover, they understand \textit{exactly} what the computer does, in such a way that at each point of the computer’s operation, they can (in principle) truly say something like ‘the computer is now considering basic map \#734, and is at such-and-such a state of checking for a 4-coloring, just as we might.’ We can capture this aspect of the operation of Appel and Haken's program by saying that it is \textit{mathematically transparent} to them.

Because Appel and Haken's program is simply performing their reasoning for them in this way, Burge thinks of Appel and Haken’s computer as a rational resource. From the fact that resources for rationality are (other things equal) to be believed, Burge concludes that `\textit{we have apriori prima facie entitlement to accept intelligible presentations-as-true, expressed by the print-outs} (of Appel and Haken's program).' \cite[p.13]{burge1998computer} By this, Burge means that Appel and Haken
have defeasible, apriori entitlement to accept as correct the output of their program, and thus 
apriori grounds to believe the 4CT. 

This warrant does not depend on an empirical fact like \textbf{Rel} about the reliability of computers. Appel and Haken infer the 4CT from the reasoning that the computer has done for them. This reasoning involves only purely mathematical considerations and not \textbf{Rel}. Nevertheless, the warrant Appel and Haken have for believing the 4CT is defeasible. For example, Appel and Haken could come to learn that the computer was malfunctioning, in which case they would no longer be justified in believing the 4CT (without further reasons). This, however, does \textit{not} mean that being justified in believing the 4CT first requires that they have positive empirical grounds for thinking that the computer was not malfunctioning. Instead, they are entitled to believe the results of mathematically transparent processes so long as they \textit{lack} reason for thinking the relevant resource unreliable. So, it is enough that Appel and Haken had no reason to think that their computer was malfunctioning.

It is also true that in making the computer and writing the program, Appel, Haken, and others had to perform all sorts of tests to establish that the computer and program were acting as they were supposed to. This, too, does not mean that the ultimate warrant for believing the 4CT is merely empirical. Empirical tests are necessary to establish that the computer is really a device that is capable of performing our reasoning for us through mathematically transparent processes. Nevertheless, once we are confident of this and use the computer in the way Appel and Haken did, the ultimate ground for accepting the 4CT is then simply the existence of a mathematically transparent process demonstrating it. This warrant is apriori insofar as it is nothing other than a mobilization of human mathematical capacities, albeit in a way that relies heavily on rational resources.

In sum, in precisely the same way that ordinary mathematicians infer theorems from the existence of purely mathematical arguments \textit{not} involving claims about human memory (even though they had to rely on their memories in convincing themselves of the existence of such arguments), so too Appel and Haken infer the 4CT from the existence of a purely mathematical argument that does \textit{not} make any claim about computers such as \textbf{Rel} (even though they had to rely on computers in convincing themselves of the existence of such an argument).

For a useful contrast, Burge considers the case of a mathematical genius whose explanations are so opaque to us that we have no real understanding of their mathematical reasoning. Insofar as their reasoning is not mathematically transparent to us, Burge thinks that we do not have the type of apriori warrant for believing them that Appel and Haken have for believing the output of their computer. Of course, if the genius's track record is sufficiently strong, we can have an \textit{inductive} warrant for believing them. We do not think, however, that this amounts to mathematical knowledge. Because of the absence of mathematical transparency, we, therefore, do not acquire mathematical knowledge from the reports of this sort of genius.

\section{Transparency and AI Assisted Proof}
\label{sec:3}

So far, we have been talking about old-fashioned computing. More recently, mathematicians have turned to deep learning models (DLMs) for assistance with particularly challenging mathematical problems in, for instance, low-dimensional topology \cite{davies2021advancing}, geometry \cite{trinh2024solving}, and extremal combinatorics \cite{romera2023mathematical}. One might expect that, like Appel and Haken, these mathematicians, too, can gain apriori knowledge from the outputs of their computers in the right circumstances.
However, the notorious opacity of deep learning machines creates a significant difference between the use of traditional computers in mathematics and the use of DLMs in mathematics.

To see this, we need to consider the sense in which DLMs are opaque and, as such, are not mathematically transparent to us. Creel's account of algorithmic and structural transparency in complex computational systems \cite{creel2020transparency} is particularly helpful in this regard.

For Creel, computational systems can be `algorithmically' and `structurally' transparent.\footnote{Creel's treatment of computational transparency can be seen as a refinement and extension of computational concepts of understanding going back to Marr \cite{marr2010vision}.} A computational system is said to be algorithmically transparent to the extent that the procedures that govern its behavior are known and intelligible. In the case of the procedures performed in the proof of the 4CT, the rules at the algorithmic level are just the rules that describe how a mathematician might check the 4-colorability of Appel and Haken's basic 1,834 graphs.   

The system is structurally transparent to the extent that it is possible to see how this algorithm is realized in actual code. Thus, a program is structurally transparent if and only if its code is surveyable, and it is possible to understand how the code generates results in accordance with the algorithm it instantiates. In cases where a computational system is algorithmically and structurally transparent (as in Appel and Haken's program), the reliability of the computational system at run-time can be (defeasibly) trusted \cite{frigg2009philosophy,duede2022instruments} (even though in practice one cannot transparently inspect it \cite{humphreys2009philosophical}).

While the computations used in Appel and Haken's program resulted in an unsurveyable proof, the computations themselves were, on Creel's account, fully algorithmically and structurally transparent. Thus the computations were mathematically transparent in the sense discussed in the last section. However, we will argue that the use of both DNNs and LLMs in mathematics are often neither structurally nor algorithmically transparent.

\subsection{Opacity of Deep Learning}
It is often said that DNNs lack epistemic transparency. It is important, however, to distinguish the \textit{training} of a DNN from fully \textit{trained} models. The procedure for training a DNN is algorithmically and structurally transparent. In most simple cases, it is algorithmically transparent that training works through the minimization of various loss functions through the iterative updating of weights on the connections between parameters in the model by means of the backpropagation of error gradients. There are extensive, well-maintained repositories that contain various structurally transparent implementations for training a wide variety of network architectures, and students taking a first class in machine learning are often required to write their own implementations of, say, the backpropagation algorithm (which, itself, involves a structurally transparent implementation of the chain rule for calculating the derivative of composed functions). There is nothing mysterious or opaque about \textit{how} users go about training such models.

However, fully \textit{trained} DNNs are said to be epistemically opaque \cite{humphreys2004extending,boge2022two}, meaning that all of the epistemically relevant factors governing the model's behavior are fundamentally unsurveyable. In general, it is not possible to determine or `fathom' \cite{zerilli2022explaining} in any intelligible or meaningful sense the algorithmic rules or principles governing the transformation of inputs to outputs of the model. As such, DNNs are opaque at the algorithmic level and, thus, at the structural level, as well. This lack of transparency is due, in large part, to the extremely high dimensionality and nonlinearity of the model, as well as the autonomous, error-driven, and semi-stochastic processes of weight assignment in guiding the settlement of the final parameterization of the model.

Of course, it is possible that with a DNN that determines whether Appel and Haken's basic 1,834 graphs are 4-colorable, we might be able to say at any moment which graph is being analyzed. There is also a numerically trivial sense in which the trained model is transparent insofar as the values on the weights themselves are available to inspection (though not surveyable) \cite{lipton2018mythos,duede2023deep}. 
However, unlike Appel and Haken's program, we would not generally be able to say \textit{how} that graph is being evaluated, and, thus, such a DNN  would neither be algorithmically nor structurally transparent.

\subsection{Mathematical Knowledge with DNNs}
Suppose that, as in the case of the 4CT, a mathematician wanted to know whether every graph in a set of graphs is 4-colorable. Approaching this problem with a DNN, the mathematician trains a model on a large set of graphs known to be 4-colorable and a large set of graphs known not to be 4-colorable. The model is then evaluated (in the usual way) on a collection of graphs not included in the training set. It is shown to correctly classify these graphs as 4-colorable or not 4-colorable and has, in fact, never misclassified the 4-colorability of any graph.

At this point, the mathematician unleashes their model on Appel and Haken's 1,834 basic graphs. After a minute or so, the model returns a result indicating that they are all  4-colorable. However, given that the model is not algorithmically transparent, we cannot regard this machine as having performed any sort of reasoning for us because we do not know what algorithm it is performing at all. Thus, the system is not mathematically transparent, and so we are not justified in believing its output on purely rational grounds. Its output, therefore, fails to give us an apriori warrant for believing the 4CT. 

However, because the DNN reliably classifies graphs as 4-colorable or not, we do get strong, inductively justified belief in the 4CT. Such a result bears some resemblance to a case \cite{davies2021advancing} considered by Duede \cite{duede2023deep}, where mathematicians use a DNN to guide mathematical attention to promising connections that led to the formulation and proof of a theorem linking specific algebraic and geometric properties of low-dimensional knots. Cases of this kind exemplify the potential for AI to assist mathematicians in their search for the most promising conjectures but leave the actual proof of the conjectures to humans. In this case our knowledge is not a direct result of the DNN, insofar as the ultimate responsibility for the proof lies with the human mathematician.

Consider, however, a hypothetical case in which a DNN trained to classify graph 4-colorability classifies all but one of Haken and Appel's 1,834 graphs as 4-colorable, except for one which it classifies as not 4-colorable. Here, the model has suggested a counterexample to the 4CT. Let us suppose that whether it is a genuine counterexample is something we can check by hand, that we check it, and we find that it is indeed a counterexample. In this case, too, we now have genuine mathematical knowledge of a mathematical fact (namely, the falsehood of the 4CT), even though, again, this knowledge cannot be said to follow \textit{directly} from the output of the DNN, as it required human verification.

\subsection{Mathematical Knowledge with LLMs}

LLMs are particularly useful for mathematics as they output reports that are potentially linguistically and mathematically intelligible. However, like DNNs, LLMs are algorithmically and structurally opaque, and so they are afflicted by the epistemic limitations discussed in the previous section.

A recent case leveraging LLMs to achieve mathematical breakthroughs in extremal combinatorics involves a treatment of the Cap Set Problem in \cite{romera2023mathematical}. A cap set is a subset of  $(\mathbb{Z} / 3\mathbb{Z})^n$ for which no three distinct elements sum to 0 (mod 3). For each $n$, the problem is to determine the size of the largest cap set. It is known that this number must be less than $\leq 3^n$ \cite{grochow19}, but its exact value is only known for $n \leq 6$. Moreover, the complexity of the solution space explodes for greater values of $n$, so brute-force computational approaches are not feasible.

In \cite{romera2023mathematical}, researchers leverage an LLM to construct a cap set of size 512 for the case $n=8$, a result that is significantly greater than the previously known largest value of 496. The approach begins by specifying an evaluation function that scores a candidate solution, where a solution is actually itself a Python program for generating a potential capset. The LLM then outputs a candidate Python program that is executed and scored by the evaluation function. If the program executes sufficiently quickly and without obvious error, it is sent to a program database. The system then samples the database and passes prior output programs to the LLM as inputs to repeat the generative process. This iterative approach generatively `evolves' candidate programs. Eventually, this process identified a cap set of size 512, which human mathematicians verified to be correct.

Unlike in the Four Color Theorem, the solution-generating procedure used here is not mathematically transparent. However, a human mathematician can easily survey and check its output. So, in this case, we get apriori knowledge that for $n=8$ there is a cap set of size 512. Nevertheless, as this involves a human mathematician verifying this fact, we cannot say that genuine mathematical knowledge has been obtained directly from the output of the computer.

\subsection{Main Claim} 
\label{sec:claim1}

With such examples in mind, we offer the following as a response to our Main Question posed in ~\ref{sec:intro}.

\begin{quote}
\textbf{Main Claim 1}: If we want to acquire apriori mathematical knowledge directly from the output of a computer, then what the computer is doing must be mathematically transparent to us (as in the case of the 4CT). If what a computer is doing is \textit{not} mathematically transparent to us (as in the case of typical DNNs or LLMs) then we cannot directly acquire apriori mathematical knowledge from the output of a computer, even though we may be able to gain a type of inductively justified belief from it. However, even if we do not directly acquire apriori mathematical knowledge from the output of a computer, if the computer outputs a human-checkable proof, example, or counterexample, then upon checking it appropriately, we do gain apriori mathematical knowledge (though not directly from the output of the computer, insofar as human checking was required.)
\end{quote}

\section{Transparent Proof Checking}
\label{sec:4}

The considerations of the previous section might be taken to entail that, while extraordinarily useful, DNNs and LMMs can ultimately only be of limited use in the acquisition of apriori mathematical knowledge. In general, their reports offer us inductively justified beliefs at best, and it is only when they output results that can be human-checked that we can acquire apriori mathematical knowledge from them.

However, there is a fairly straightforward way to surpass these limits in certain cases. Let us focus on the case in which a machine (perhaps an LLM) outputs not only some mathematical claim $X$ but also something it claims to be a proof of $X$, and that this proof is stored somewhere on a hard drive. If what has been stored on the drive can be human-checked, then we can check it, and if it is indeed a correct proof, we thereby gain apriori knowledge of $X$. 

Suppose, however, that the proof of $X$ stored on the hard drive is so long that it cannot be human-checked. It might then appear that apriori knowledge of $X$ is beyond our reach.

But this is not so. Let us suppose that the stored proof of $X$, while enormously complex, is systematically organized as a sequence or tree of propositions of the sort one might encounter in a mathematical logic class. We can imagine constraining the output of the machine generating the proof in such a way as to demand that its outputs be formulated in this way (as in \cite{romera2023mathematical} where the model outputs all results in syntactically correct \texttt{Python} or, as is increasingly common \cite{avigad2024mathematics}, in a formal language instantiated in something like \texttt{Lean} \cite{de2015lean}). We can even allow various abbreviations, additional rules, and verbose articulations of the steps in the proof so that this proof has roughly the form of a human-generated proof,\footnote{Of course, the proofs of practicing mathematicians are not like this, often involving large leaps and a kind of hand-wavy skipping over of `trivial' steps \cite{kitcher1998mathematical}.} even though it was not generated by a human and is in fact so large that it cannot be surveyed by a human. These assumptions can be made to hold by adding some overhead to the original program and forcing the proof of $X$ to be stored in this form.

Although we cannot check this proof ourselves, this does not stop us from writing a proof-checking program that can. The proof-checking program we are imagining goes through the proof, verifying that it starts with genuine axioms and that each step is a legitimate application of some standard logical or mathematical rule. We can imagine a version of this proof-checker that is, in fact, completely mathematically transparent, approaching the task of checking the proof in exactly the way a human would. When such a program runs, at any point, we can (in principle) correctly say something like `the computer is now checking inference 15435 and is verifying that it is a correct application of modus ponens.'

Let us assume that we run this proof-checker, and it reports no errors. Just as Appel and Haken acquire apriori knowledge of the 4CT from the output of their mathematically transparent program, so too we acquire apriori knowledge that there is a correct proof of $X$ from the output of our mathematically transparent proof-checker. From the fact that there is a correct proof of $X$, the truth of $X$ follows, and thus we acquire apriori knowledge of $X$. This is true even though no human has (or ever could have) any sort of rational grasp on the process that led to the generation of the proof, and no human is capable of checking the proof stored on the hard drive.

The important point, however, is that in this case, we have apriori knowledge of $X$ not based on the output of the LLM, whose workings are not transparent to us, but based on the output of the proof-checker, whose workings are transparent to us.

Of course, if the LLM `claims' to have proven $X$ but cannot produce and store the actual proof of $X$, then we cannot use a proof-checker in the way just described. In this case, we see no way to acquire anything other than inductive grounds for believing $X$. 

This leads us to the second of the two central claims of this paper, which may be viewed as a counterpoint to Main Claim 1.

\begin{quote}
\textbf{Main Claim 2}:  We \textit{can} (indirectly) gain apriori knowledge from the output of a computer program that is not mathematically transparent but which stores a (not necessarily human-checkable) proof of a mathematical claim. This is accomplished by employing a mathematically transparent proof-checker to evaluate the stored proof of the claim.
\end{quote}

\section{General Conclusions}
\label{sec:5}

Modern LLMs and DNNs are opaque to us in ways that create obstacles to obtaining mathematical knowledge from them. However, we have argued that if a proof-checker, transparently automating human forms of mathematical evaluation, is attached to such machines, then we can obtain apriori mathematical knowledge from them. Surprisingly, this applies even in cases where the original machines are entirely opaque to us and the proofs they output are not human-surveyable.

A different question for further consideration is to what extent we may gain scientific \cite{birhane2023science} knowledge outside of mathematics by appending analogous transparent `checking' mechanisms to the output of otherwise opaque algorithms. This would get us closer to overcoming the perceived problems of confabulation and realizing the ambition of fully automated scientific discovery.

\bibliographystyle{alpha}
\bibliography{bibliography}
 
\end{document}